\documentclass[manuscript,screen]{acmart}
\usepackage{array}

\AtBeginDocument{%
  \providecommand\BibTeX{{%
    \normalfont B\kern-0.5em{\scshape i\kern-0.25em b}\kern-0.8em\TeX}}}

\setcopyright{acmcopyright}
\copyrightyear{2023}
\acmYear{2023}
\acmDOI{XXXXXXX.XXXXXXX}

\acmConference[OzCHI'23]{Make sure to enter the correct
  conference title from your rights confirmation emai}{December 02--06,
  2023}{Wellington, NZ}
\acmBooktitle{OzCHI'23: ACM Conference of the Computer-Human Interaction Special Interest Group of Australia on Computer-Human Interaction,
 December 03-06, 2023, Wellington, NZ} 
\acmPrice{15.00}
\acmISBN{978-1-4503-XXXX-X/18/06}

\begin{document}


\title[Contextually-Adaptive Human-Robot Interactions in Urban Public Environments]{Robots in the Wild: Contextually-Adaptive Human-Robot Interactions in Urban Public Environments}

\author{Xinyan Yu}
\authornote{Both authors contributed equally to this research.}
\email{xinyan.yu@sydney.edu.au}
\orcid{0000-0001-8299-3381}
\affiliation{Design Lab,
  \institution{The University of Sydney}
  \city{Sydney}
  \state{NSW}
  \country{Australia}
}

\author{Yiyuan Wang}
\email{yiyuan.wang@sydney.comm.au}
\orcid{0000-0003-2610-1283}
\authornotemark[1]
\affiliation{Design Lab,
  \institution{The University of Sydney}
  \city{Sydney}
  \state{NSW}
  \country{Australia}
}

\author{Tram Thi Minh Tran}
\email{tram.tran@sydney.edu.au}
\orcid{0000-0002-4958-2465}
\affiliation{Design Lab,
  \institution{The University of Sydney}
  \city{Sydney}
  \state{NSW}
  \country{Australia}
}

\author{Yi Zhao}
\email{yi.zhao1@sydney.edu.au}
\orcid{0000-0003-2803-0933}
\affiliation{Design Lab,
  \institution{The University of Sydney}
  \city{Sydney}
  \state{NSW}
  \country{Australia}
}

\author{Julie Stephany Berrio Perez}
\email{stephany.berrioperez@sydney.edu.au}
\orcid{0000-0003-3126-7042}
\affiliation{The Australian Centre for Robotics,
  \institution{The University of Sydney}
  \city{Sydney}
  \state{NSW}
  \country{Australia}
}

\author{Marius Hoggenmüller}
\email{marius.hoggenmuller@sydney.edu.au}
\orcid{0000-0002-8893-5729}
\affiliation{Design Lab,
  \institution{The University of Sydney}
  \city{Sydney}
  \state{NSW}
  \country{Australia}
}

\author{Justine Humphry}
\email{justine.humphry@sydney.edu.au}
\orcid{0000-0002-2376-2089}
\affiliation{Faculty of Arts and Social Sciences,
  \institution{The University of Sydney}
  \city{Sydney}
  \state{NSW}
  \country{Australia}
}

\author{Lian Loke}
\email{lian.loke@sydney.edu.au}
\orcid{0000-0001-7174-8209}
\affiliation{Design Lab,
  \institution{The University of Sydney}
  \city{Sydney}
  \state{NSW}
  \country{Australia}
}
\author{Lynn Masuda}
\email{rin.masuda@sydney.edu.au}
\orcid{0009-0002-4989-1126}
\affiliation{Design Modelling \& Fabrication (DMaF) Lab,
  \institution{The University of Sydney}
  \city{Sydney}
  \state{NSW}
  \country{Australia}
}

\author{Callum Paker}
\email{callum.parker@sydney.edu.au}
\orcid{0000-0002-2173-9213}
\affiliation{Design Lab,
  \institution{The University of Sydney}
  \city{Sydney}
  \state{NSW}
  \country{Australia}
}

\author{Martin Tomitsch}
\email{Martin.Tomitsch@uts.edu.au}
\orcid{0000-0003-1998-2975}
\affiliation{Transdisciplinary School,
  \institution{University of Technology Sydney}
  \city{Sydney}
  \state{NSW}
  \country{Australia}
}

\author{Stewart Worrall}
\email{stewart.worrall@sydney.edu.au}
\orcid{0000-0001-7940-4742}
\affiliation{The Australian Centre for Robotics,
  \institution{The University of Sydney}
  \city{Sydney}
  \state{NSW}
  \country{Australia}
}

\renewcommand{\shortauthors}{Yu and Wang, et al.}

\begin{abstract}

The increasing transition of human-robot interaction (HRI) context from controlled settings to dynamic, real-world public environments calls for enhanced adaptability in robotic systems. This can go beyond algorithmic navigation or traditional HRI strategies in structured settings, requiring the ability to navigate complex public urban systems containing multifaceted dynamics and various socio-technical needs. Therefore, our proposed workshop seeks to extend the boundaries of adaptive HRI research beyond predictable, semi-structured contexts and highlight opportunities for adaptable robot interactions in urban public environments. This half-day workshop aims to explore design opportunities and challenges in creating contextually-adaptive HRI within these spaces and establish a network of interested parties within the OzCHI research community. By fostering ongoing discussions, sharing of insights, and collaborations, we aim to catalyse future research that empowers robots to navigate the inherent uncertainties and complexities of real-world public interactions.
\end{abstract}

\ccsdesc[500]{Human-centered computing~Interaction design}

\keywords{urban robots, human-robot interaction, adaptive interaction}



\maketitle

\section{Introduction}
Rapid advancements of Internet of Things (IoT) and artificial intelligence (AI) technologies have transformed robots from being mere task executors to context-aware entities capable of exchanging information and making autonomous decisions~\cite{vermesan2017internet}. As a consequence, the context of human-robot interaction (HRI) is witnessing a shift from highly controlled settings (e.g. manufacturing sites where robots strictly follow pre-programmed instructions), to more dynamic and unstructured environments that require flexibility for achieving better interaction outcomes (e.g. improved user experience~\cite{tielman2014adaptive,cumbal2022shaping}, enhanced task completion ~\cite{bajones2016enabling,de2018effect}). This transition encompasses HRI in a variety of less-structured settings, ranging from service robots in indoor shared spaces like shopping malls~\cite{Kanda2009Guide} and airports ~\cite{Nielsen2018Airport, Joosse2017Airport}, to autonomous vehicles in outdoor public areas such as recreational zones ~\cite{hoggenmueller2020stop} and urban traffic environments~\cite{Yu2023Prediction,wang2022pedestrian}. These diverse and dynamic scenarios introduce new challenges for robots' operations and require them to adapt to the inherent uncertainties and complexities of real-world interactions with humans.



Adaptability in HRI has been an area of significant interest and investigation. Current endeavours primarily focus on semi-structured settings, such as workplaces or domestic environments. For example, social robots can sense the psychological and emotional states of their human interaction partners and react accordingly~\cite{Gena2022Wolly,zedda2020designing,tielman2014adaptive,cumbal2022shaping}, making interactions natural and engaging. In task-specific situations, robots can assess the level of task knowledge possessed by human partners and adjust communication strategies to improve efficiency and user experience, such as in educational \cite{de2018effect,cumbal2022adaptive} or human-robot collaboration (HRC) ~\cite{Kontogiorgos2020LeastEffort, Torrey2007AdaptiveDialogue} scenarios. In addition to recognising various internal states of users, robots have been tailored to diverse user demographics, such as age, gender, and health conditions \cite{Younis2023Age,canete2021pepe,bajones2016enabling}, and conquering limitations of physical environments, such as walls and rugged terrains \cite{wu2020design,kim2011terrain}.
While extensive research has delved into the adaptability of HRI in semi-controlled settings, where interactions with authentic users (i.e. people who have the intent to interact with the robot ~\cite{Nielsen2018Airport}) tend to be more predictable~\cite{Putten2020Forgortten}, explorations for those robots in the wild are currently limited to technical developments, e.g. algorithmic navigation of autonomous robots \cite{li2020socially,chen2017socially}. The interaction strategies that various robots could employ while operating in complex, urban public areas and the transferability of existing strategies to such contexts remain under-explored. 

Unlike robots operating in relatively static environments, which interact with a consistent group of users, adaptability for robots in public urban environments implies a much broader scope beyond just adjusting to an individual's state or demographic differences.
For example, individuals who casually encounter robots in public spaces are typically not expected to interact with them ~\cite{Putten2020Forgortten}, as these non-users may already be occupied with other tasks (e.g., people engaging with their phones are potentially unaware of and less likely to interact with robots in their surroundings). 
In addition to adapting to individual characteristics, people in public spaces can often engage in social interactions with others in groups, which may require distinct interaction strategies tailored to collective patterns. For instance, observations in the wild found that pedestrians tend to follow others' behaviours, such as crossing the road in crowds or gazing in the same direction \cite{wang2022pedestrian,faria2010collective,gallup2012visual}. Acquainted groups can demonstrate social dynamics (e.g. chatting) which robots should respect, such as by not navigating into the group and maintaining appropriate proxemics \cite{wang2022pedestrian,li2020socially,chen2017socially}. From a broader perspective, when robots are interactive urban applications embedded in the real world \cite{hoggenmueller2020stop}, their stakeholders can transcend individual entities and further involve the local community as a whole~\cite{Tomitsch+2017}, which necessitates higher-level considerations around the dynamics and needs of the community they serve. At the same time, HRI in the wild can be impacted by many factors stemming from natural or built environments, including a wide range of elements such as weather conditions, road infrastructure, vegetation characteristics, and the coexistence of diverse animal species. Consequently, these factors pose complex challenges for robots' deployment in public settings and adapting to the contextual influences during interactions with human counterparts.




\section{Workshop Objectives}


In response, this workshop seeks to broaden the scope of adaptive HRI research which by far mostly serves predictable settings, through adopting a more holistic perspective around real-world challenges and highlighting opportunities for adaptable robot interactions in urban public environments. The half-day workshop will focus on two main objectives: (1) exploring design opportunities and challenges in contextually-adaptive HRI within urban public spaces, and (2) establishing a network of interested parties within the OzCHI research community to facilitate ongoing discussions, sharing of insights, and collaborations in this field.

\section{Workshop Overview}

\subsection{Pre-Workshop Plan}

We have planned to develop a website dedicated to providing detailed information about the workshop, offering a centralised hub for disseminating information and updates. To ensure we reach a broad audience, we will share the workshop's call for participants across social media platforms like Twitter and LinkedIn, and leverage our professional networks. 
To ensure our event runs smoothly and meets our objectives, the organising team will pilot the procedures and activities prior to the workshop day.  

\begin{figure}[h]
\begin{center}
\includegraphics[width=0.5\textwidth]{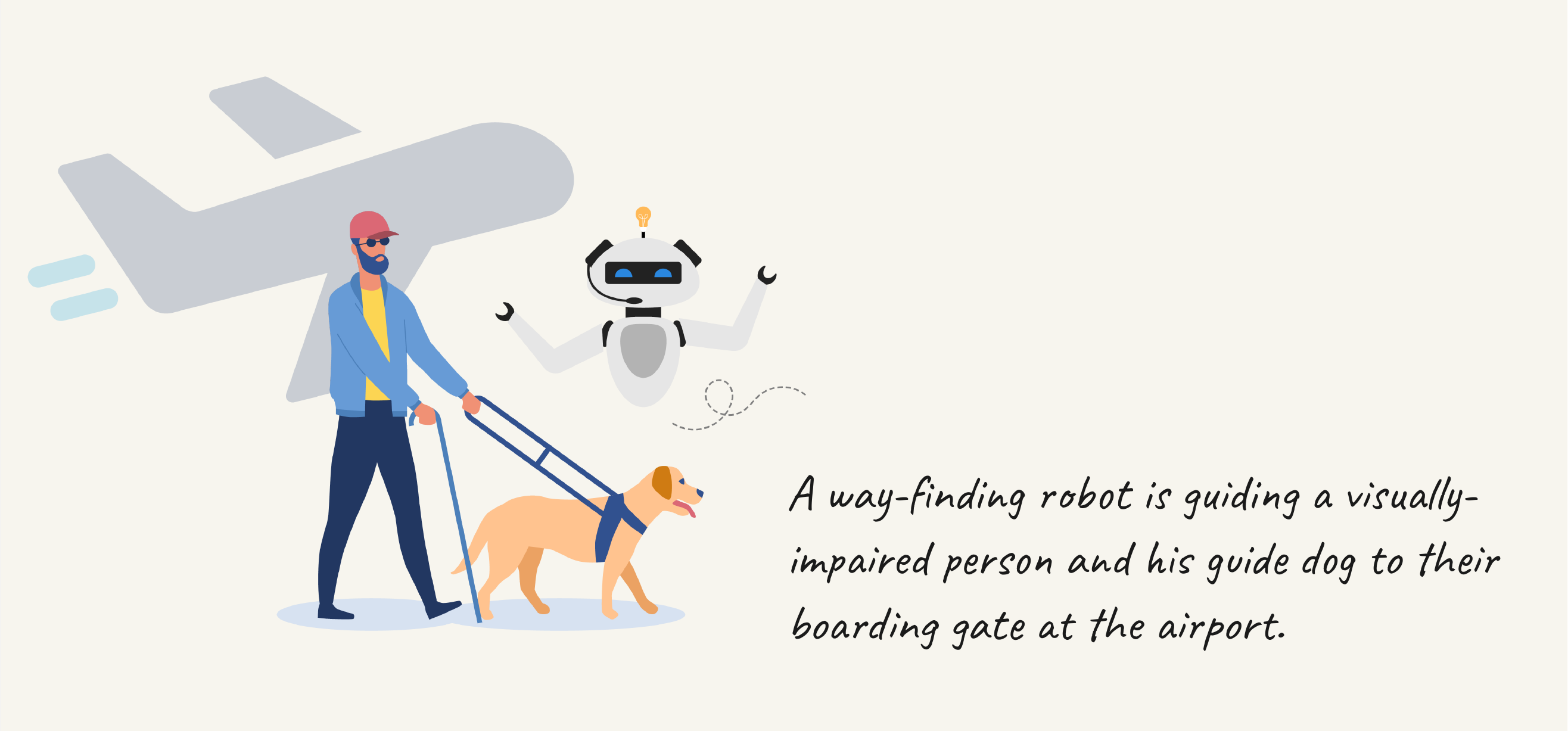}
\end{center}
\caption{Example of a contextually-adaptive HRI scenario: Way-finding robots working at the airport adapting to varied demographics and accessibility needs, e.g., aiding a visually impaired person with a guide dog.}
\label{fig:example}
\Description{}
\end{figure}

In preparation for the workshop, we invite participants to envision and speculate on a specific scenario where the interaction between humans and robots needs to adapt to the context. Participants can encapsulate this scenario visually through a photograph or a simple sketch. If necessary, a brief textual explanation can accompany this visual to elaborate further and provide additional clarity (refer to Figure \ref{fig:example} for an example). This visual representation helps stimulate their thinking around the topic, leading to richer and more informed discussions. Participants will need to submit the prepared material before the workshop and carry a printed copy to the event.
\subsection{Workshop Activities}

\begin{table}[h]
  \small
  \caption{Tentative schedule activities of the workshop}
  \label{tab:schedule}
  \begin{tabular}{l | l | l}
    \toprule
    \textbf{Activity} &\textbf{Duration} & \textbf{Time} \\
    \midrule
    Opening \& Introduction & 15 min & 9:00 - 9:15\\ 
    \textbf{Keynote Speak} & 30 min & 9:15 - 9:45 \\ 
    \textbf{Urban Context Reflective Activity } & 60 min & 9:45 - 10:45\\ 
    Coffee Break & 15 min & 10:45 - 11:00 \\
    \textbf{Presentation \& Discussion} & 45 min & 11:00 - 11:45 \\ 
    Wrapping-up \& Next Steps & 30 min &  11:45 - 12:15 \\
  \bottomrule
\end{tabular}
\end{table}

The workshop will be held in person, with an estimated attendance of 10-30 participants. During the workshop, participants will engage in a city walk reflective activity facilitated by a cultural probe ~\cite{van2016eliciting} developed by us and lively discussions. 

\subsubsection{Invited Keynote Speaker}
Maria Luce Lupetti will be the keynote speaker at the workshop. She is an Assistant Professor in Interaction and Critical Design at the faculty of Industrial Design Engineering, TU Delft (NL). Her research, at the intersection of design~\cite{Lupetti2021Designerly, Lupetti2018fiction}, ethics~\cite{cavalcante2023meaningful}, AI~\cite{murray2023grasping,Marius2023CreativeAI} and robotics ~\cite{Lupetti2017Children, Lupetti2023Embodied}, is focused on promoting critical perspectives in technology development. She is also a core member of the AiTech Initiative on Meaningful Human Control over AI Systems, a member of the Dutch Social Justice and AI network, and serves as Exhibit X section editor for Interactions Mag.

\subsubsection{Urban Context Reflective Activity}

\begin{figure}[h]
\begin{center}
\includegraphics[width=0.5\textwidth]{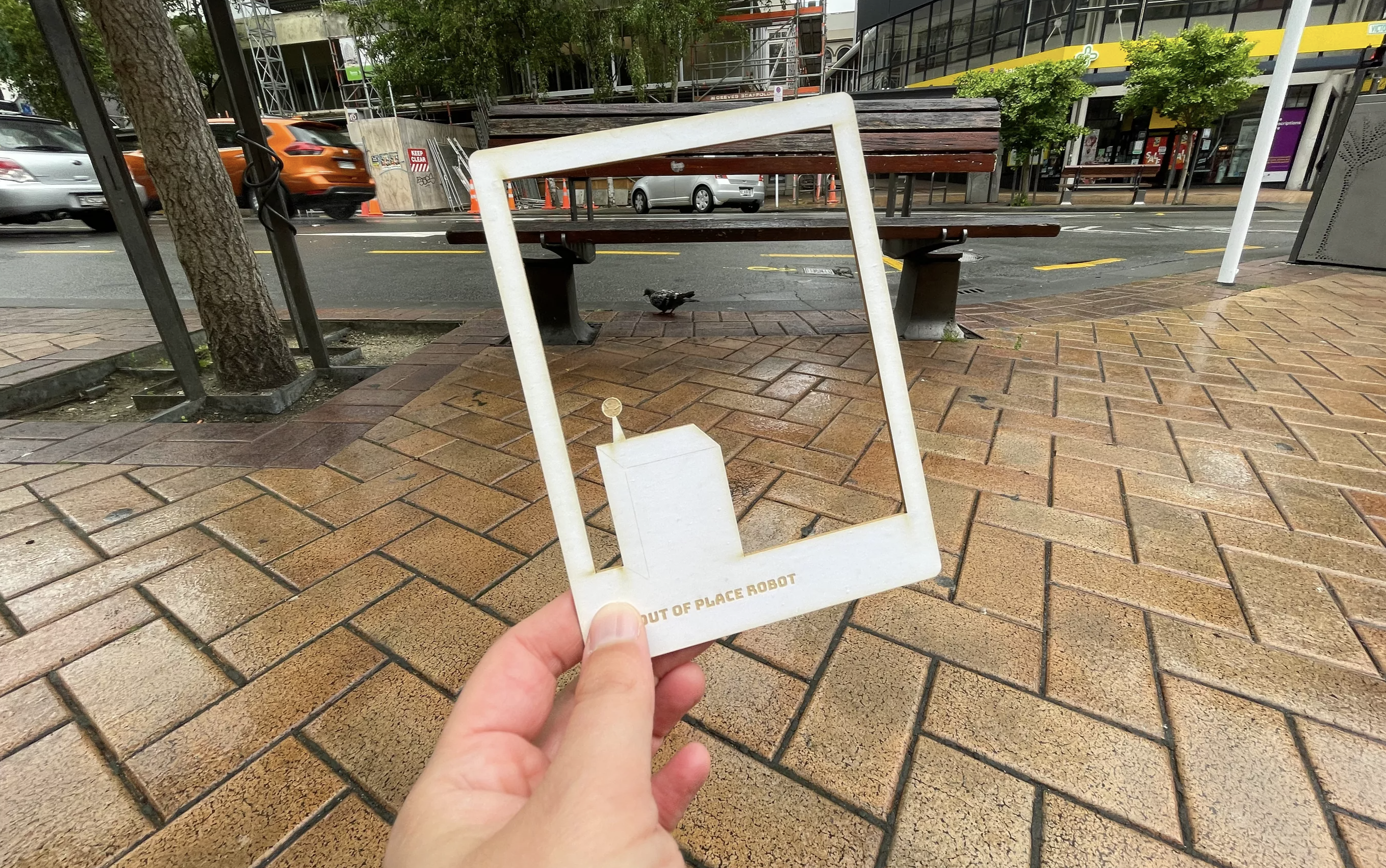}
\end{center}
\caption{Example of taking a photo with the probe}\label{probe}
\Description{}
\end{figure}

To investigate real-world contexts where robots might be misfits and to gain preliminary insights into their contextual adaptability, we developed a cultural probe~\cite{Gaver1999CulturalProbes}. The cultural probe we designed was a pocket-sized photo frame (see Fig. ~\ref{probe}), featuring an abstract image of a robot. It helps to elicit imaginative reflections by conceptually integrating robots into various scenarios of everyday urban life.
Participants will be split into groups of three or four people and engage in a physical city walk around the workshop venue. Each participant will be given a cultural probe and instructed to freely walk around and use the probe to take pictures of scenarios where they feel the presence of the robot would be out of place.

The city walk activity lasted for approximately one hour. Afterwards, participants will be asked to submit their captured images, descriptions of each scenario, and their thoughts on why the robot seemed out of place, to a public content-sharing board, Padlet\footnote{Padlet https://padlet.com}.

\subsubsection{Open Discussion}
After the city walk activity, participants will be divided into two groups, with each group being facilitated by one of our organising members. All participants will present the scenarios they captured during the previous city walk. Then, each group will discuss themes emerging from these scenarios, along with challenges identified from the posts. These discussions will include topics such as (1) potential future research questions related to contextually adaptive human-robot interaction (HRI), and (2) strategies to address these challenges. At the end of this session, each group will be asked to briefly summarise their discussion.




\subsection{Post-Workshop Plan and Expected Outputs}
The workshop is expected to identify concrete design opportunities and challenges for adaptive HRI in urban public settings. Additionally, the workshop aims to raise awareness among the HRI research community about these issues and stimulate further research in this area. These valuable findings will be consolidated into principles and guidelines for contextually adaptive HRI in public settings. The outcomes of the workshop will then be shared through a related publication to ensure a broader reach and increase its impact, with all participants invited as co-authors. We hope the workshop will encourage ongoing exploration and development in the field of adaptive HRI in public settings, motivating researchers and practitioners to contribute to the advancement of this area.  

\subsection{Call for Participants}

While the workshop centres on HRI, we acknowledge the significance of interdisciplinary collaboration in addressing the challenges of contextual adaptive HRI in public urban settings due to the involvement of multifaceted contextual elements. To foster a holistic approach, this workshop will assemble academic researchers and industry practitioners from multiple disciplines such as Human-Computer Interaction, Robotics and Engineering, Transportation, Design and Urbanism, Social Sciences, Psychology, Ethics, and AI, among others. By leveraging their insights and expertise, we anticipate the development of conceptual visions that shed light on the future of HRI research and its ability to adapt to various contexts in dynamic and complex public settings.


We ask participants to prepare a visual representation that communicates their speculations on an adaptive robot scenario or any robot adaptability issue they have encountered based on past experiences. This visual representation can take the form of a photograph or a simple sketch that conveys the essence of their ideas. If needed, participants can provide a brief textual explanation to further clarify and elaborate on their visual representation (refer to Fig.\ref{fig:example} for an example). This representation will act as a medium to stimulate discussions and foster reflective thinking. Participants are encouraged to send their visual representations to the organising team prior to the workshop. 

\section{Organisers}
\textbf{Xinyan Yu} is a PhD candidate in the Design Lab at the University of Sydney's School of Architecture, Design, and Planning. Her research centres around human-robot collaboration in urban settings, exploring bystanders' pro-social behaviours towards robots.

\textbf{Yiyuan Wang} is a PhD candidate in the Design Lab at the University of Sydney's School of Architecture, Design, and Planning. Her current research focuses on designing and prototyping human-machine interfaces to support autonomous vehicle/robot-pedestrian interactions in social traffic environments.

\textbf{Tram Thi Minh Tran} is a PhD candidate in the Design Lab at the University of Sydney's School of Architecture, Design, and Planning. Her research investigates the intricate interactions between pedestrians and autonomous vehicles in complex traffic scenarios involving multiple road users. 

\textbf{Yi Zhao} is a PhD candidate in the Robotic Lab at the University of Sydney. His work focuses on constructing a new design framework for human-robot collaboration (HRC) tasks and aims at inspiring designers to explore the characteristics of robotic behaviour and interaction, enabling robots to communicate and work with humans as partners in the HRC project.

\textbf{Dr. Julie Stephany Berrio} holds the position of Research Associate at the Australian Centre for Robotics, which is affiliated with the University of Sydney. Her research is primarily centred around perception, mapping, and digital twins, specifically focusing on their application in vehicular technology.

\textbf{Dr. Marius Hoggenmüller} is a Lecturer in Interaction Design in the Design Lab at the University of Sydney's School of Architecture, Design, and Planning. His work focuses on prototyping interactions with emerging technologies in cities, such as urban robots and autonomous systems.

\textbf{Dr. Justine Humphry} is a Senior Lecturer in Digital Cultures in the Discipline of Media and Communications at the University of Sydney. Her research examines the cultures and politics of mobile and digital media in everyday life, with a focus on the digital experiences and challenges of under-represented and excluded communities.

\textbf{Dr. Lian Loke} is an Associate Professor in the Design Lab at the University of Sydney's School of Architecture, Design, and Planning. She brings a choreographic and somaesthetic approach to staging and studying interactions with people and machines in artistic, domestic and public contexts.

\textbf{Lynn Masuda} is a Robotics Prototyping Officer in the Design, Modelling and Fabrication Lab (DMaF Lab) at The University of Sydney. Her work involves training students, academic research and task programming with industrial and collaborative robotic arms for design, architecture and construction processes.

\textbf{Dr. Callum Parker} is a Lecturer in Interaction Design at the Urban Interfaces Lab in the University of Sydney’s School of Architecture, Design and Planning. His research seeks to gain new understanding of interactive digital city interfaces and their place within urban environments, contributing towards smarter cities.

\textbf{Dr. Martin Tomitsch} is a Professor and Head of the Transdisciplinary School at the University of Technology Sydney, and a founding member of the Media Architecture Institute, the Urban Interfaces Lab, and the Life-centred Design Collective. He works at the intersection of design and technology with a focus on cities and responsible innovation.

\textbf{Dr. Stewart Worrall} is a Senior Research Fellow, and leads the Intelligent Transportation Systems group at the Australian Centre for Robotics which is part of the University of Sydney. His research aims to track and predict the intentions of drivers and pedestrians, and better understand how this can be used to improve the way that vehicles interact with people.


\bibliographystyle{ACM-Reference-Format}
\bibliography{Adaptive_HRI}

\end{document}